\pdfoutput=1

\documentclass[11pt]{article}

\usepackage[review]{acl}

\usepackage{times}
\usepackage{latexsym}

\usepackage[T1]{fontenc}

\usepackage[utf8]{inputenc}

\usepackage{microtype}

\usepackage{bm}

\usepackage{amsmath}

\usepackage{graphicx}
\graphicspath{ {./images/} }

\title{Syntax-driven Data Augmentation for Named Entity Recognition}

\author{
    Arie Pratama Sutiono \\ 
    University of Arizona \\  
    Linguistics \\  
    \texttt{ariesutiono@arizona.edu} \\
    \And
    Gus Hahn-Powell \\ 
    University of Arizona \\  
    Linguistics \\  
    \texttt{hahnpowell@arizona.edu}\\}
        
\begin{document}
\maketitle
\nolinenumbers
\begin{abstract}
In low resource settings, data augmentation strategies are commonly leveraged to improve performance. Numerous approaches have attempted document-level augmentation (e.g., text classification), but few studies have explored token-level augmentation. Performed naively, data augmentation can produce semantically incongruent and ungrammatical examples. In this work, we compare simple masked language model replacement and an augmentation method using constituency tree mutations to improve the performance of named entity recognition in low-resource settings with the aim of preserving linguistic cohesion of the augmented sentences.

\end{abstract}

\section{Introduction}
Deep neural networks have proven effective for a wide variety of tasks in natural language processing; however, these networks often require large annotated datasets before they begin to outperform simpler models. Such data is not always available or sufficiently diverse, and its collection and annotation can be an expensive and slow process.  The trend of fine-tuning large-scale language models originally trained using self-supervision has helped to alleviate the need for large annotated datasets, but this approach relies on the dataset for fine-tuning being diverse enough to train a model that generalizes well.  Careful data augmentation can help to improve dataset diversity and ultimately the model's generalizability.

Data augmentation, a technique to generate data given training set characteristics, continues to play a critical role in low-resource settings; however, the majority of work on data augmentation focuses on improving document-level tasks such as text classification. Far less attention has been paid to token-level tasks~\citep{feng-etal-2021-survey}. 

Prevailing approaches to sequence tagging tasks such as named entity recognition (NER) require token-level ground truth. Naive replacement-based methods for augmentation may introduce noise in the form of sentences that are ungrammatical, semantically vacuous, or semantically incongruent. Whether in the form of insertions, deletions, or substitutions, care must be taken with token-level augmentation to preserve linguistic cohesion. 

There is evidence that large language models may possess some syntactic knowledge \citep{hewitt-manning-2019-structural,wei-etal-2021-frequency}. Work by \citet{bai-etal-2021-syntax} suggests that incorporating syntactic tasks into pre-training improves the performance of large language models. Inspired by these findings, we investigate how constituency trees might be used to guide data augmentation. We use tree-based transformations to mutate sentences while minimizing undesired side effects of syntactic manipulation (i.e., preserving linguistic cohesion). Related work by \citet{zhang-etal-2022-treemix} explores a similar approach in different settings. They focus on the effect of constituency based replacement in single classification and pair sentence classification tasks, while this work examines token-level classification.

We compare our syntax-driven method with no augmentation (our baseline), augmented data generated through cloze-style~\citep{cloze} masked language modeling using a BERT-based classifier, successful approaches introduced by \citet{Dai2021}, and two of the top-performing augmentation strategies according to past work: synonym-based replacement and mention-based replacement. Following prior work, we We use the i2b2-2010 English language dataset \cite{Uzuner20112010IC} for NER.

\section{Related work}

A variety of approaches have been explored for document-level data augmentation. The usage of backtranslation to generate augmented samples was introduced by \citet{kobayashi-2018-contextual}. \citet{wei-zou-2019-eda} explored synonym replacement, random insertion, random swap, and random deletion for text classification. \citet{quteineh-etal-2020-textual} introduced an approach using monte carlo tree search (MCTS) to guide generation of synthetic data.

Augmentation for token-level sequence tagging, however, is understudied. Simple approaches for token-level classification (e.g., synonym replacement, mention replacement, shuffling, etc.) was investigated by \citet{devlin-etal-2019-bert}. They used a sample of 50, 150, and 500 sentences to simulate a low-resource setting. Token linearization (TL) was introduced by \citet{ding-etal-2020-daga}. The main idea of TL is to incorporate NER tags inside the training sentences themselves. \citet{ding-etal-2020-daga} experimented with various sizes of training data across six languages.

A Similar idea to that presented here has been used in parallel research by \citet{zhang-etal-2022-treemix}. Their approach, TreeMix, works similarly to our approach in that it replaces a phrase with another phrase from another another training instance by swapping phrases with the same constituent labels. \citet{zhang-etal-2022-treemix} demonstrated that TreeMix outperforms a method that selects a random span of text as the replacement for a target phrase, suggesting that syntactically=aware replacement can improve data augmentation for at least some tasks. 

\section{Approach}

\subsection{Synonym Replacement (SR)}
\citet{Dai2021} experimented with replacing randomly selected tokens from the training corpus with a multiword synonyms originating from WordNet \citep{miller-1992-wordnet}. In this case, if the replaced token is the beginning of a mention (\textsc{B-ENTITY}), then the first token of the synonym will be tagged as \textsc{B-ENTITY} and the rest will be considered as \textsc{I-ENTITY}. In cases where the replaced token is in the middle of a mention (\textsc{I-ENTITY}), then all of the synonym's tokens will be assigned to \textsc{I-ENTITY}.

\subsection{Mention Replacement (MR)}
\citet{Dai2021} described mention replacement as using a Bernoulli distribution to decide whether each mention should be replaced. If yes, then another mention which has the same entity type as the target mention from the original training set is selected to replace the target mention. For instance, if the mention "myelopathy / \textsc{B-PROBLEM} is selected for replacement, then we can select one of \{"C5-6", "COPD", ...\} which all have the same entity type (\textsc{PROBLEM}).

\subsection{Language Model (LM)}
We experimented with token replacement using a masked language model. We restrict the system to replace only non-mention tokens (tokens with category \textsc{O}). This is because if we replace tokens with a named entity, we cannot guarantee that the output from the masked language model will have the same category, such that if we replace a token categorized as \textsc{B-TEST}, we could not guarantee that the masked language model will replace it with a similar token to those in \textsc{B-TEST} category.

We randomly select, without replacement, $n$ tokens as candidates to be replaced. The selected tokens are masked from the original sentence.  Next, we the language model generates replacements for the masked tokens. We may repeat this token generation up to $k$ times to generate different augmented sentences. We use Allen AI's SciBERT model from the Hugging Face model repository. 

\subsection{Constituency Replacement (CR)}
As a preprocessing step, we perform constituency parsing over all of the training data using Stanza~\citep{DBLP:journals/corr/abs-2003-07082}.  Given an XP non-terminal, we select $p$ non-terminals as candidates for replacement. For each non-terminal, we find other non-terminals with the same category from the training data, to replace the candidate. Assuming that we chose VP as the non-terminal taget node for replacement, the algorithm will choose another VP from the set of parsed sentences in the training corpus and mutate the whole subtree (VP root and the nodes below it). We can repeat this process to generate $k$ augmented sentences. Additionally, we target nodes that have NER mentions as one of its children.

\section{Experiments and Results}

\subsection{Dataset}
We used the i2b2-2010 dataset \citep{Uzuner20112010IC}, an English language NER dataset. Similar to \citet{Dai2021} we use 3 different sizes of dataset to simulate low-resource settings. We select the first 50, 150, and 500 sentences from training set and denote them as S, M, L. We used the default train-test split and limit the augmentations to training set.

\subsection{Model}
Following \citet{Dai2021}, we model NER task as sequence-labeling. We used same components for modeling: a neural encoder and a conditional random field layer. For our neural encoder, We used SciBERT. This model has been proven to work effectively with scientific and medical data, like i2b2-2010. 

\subsection{Experiments}

Each experiment was repeated with 5 different random seeds to calculate standard deviation.

For our SR and MR approaches, 
our hyperparameters are the replace ratio (0.3) and the number of generated samples (1). 

LM and CR hyperparameters are similar. Both will have the number of generated samples and number of replaced tokens (only for LM) or non-terminals (only for ST). 
In this work, we limit CR replacements to non-terminals (phrases) in the following set: \{NP, VP, ADJP, ADVP, PP\}. We leave out FRAG (fragment) because it has too low of a non-terminal count.

One question to be explored is whether more augmented data can result in continued gains in model performance. To answer this, we experimented with \{5, 10, 20\} generated samples. For each number of generated samples, we also set the number of replaced tokens for the LM and the number of replaced non-terminals for CR to be \{1, 3, 5\}. 
We have described the distribution of non terminals in Table~\ref{tab:nonterminalcounts}. All of these settings were tested against the 27,625 sentences from our validation set. 

\begin{table}[h]
\centering
\begin{tabular}{llll}
\textbf{Phrase} & \textbf{S} & \textbf{M} & \textbf{L} \\
\hline
{NP} & 332 & 637 & 2562 \\
{VP} & 93 & 189 & 881 \\
{PP} & 54 & 130 & 690 \\
{ADJP} & 31 & 42 & 189 \\
{ADVP} & 16 & 27 & 126 \\
{FRAG} & 2 & 2 & 4 \\
\hline
\end{tabular}
\caption{\label{tab:nonterminalcounts} Distribution of the number of phrases in the training data.}
\end{table}

\subsection{Results}
Table~\ref{tab:f1results} described the highest F1 scores for each augmentation strategy. The best F1 scores were taken for each strategy, across multiple hyperparameters. We found that synonym replacement still outperforms other augmentation strategies in small and medium dataset sizes.

\begin{table}[h]
\centering
\scalebox{.8}{%
\begin{tabular}{llll}
\textbf{Experiment} & \textbf{S} & \textbf{M} & \textbf{L} \\
\hline
{NoA} & {$46.3 \pm 0.5$} & {$61.4 \pm 0.1$} & {$70.7 \pm 0.1$} \\
{SR} & {$\bm{53.0 \pm 0.2}$} & {$\bm{65.7 \pm 0.1}$} & {$71.0 \pm 0.0$} \\
{MR} & {$51.9 \pm 0.2$} & {$61.7 \pm 0.1$} & {$70.2 \pm 0.0$} \\
{LM} & {$52.9 \pm 0.1$} & {$63.3 \pm 0.1$} & {$\bm{73.3 \pm 0.2}$} \\
{CR-ADJP} & {$47.8 \pm 0.2$} & {$61.0 \pm 0.1$} & {$71.2 \pm 0.1$} \\
{CR-ADVP} & {$50.5 \pm 0.3$} & {$61.9 \pm 0.1$} & {$71.3 \pm 0.1$} \\
{CR-NP} & {$52.1 \pm 0.3$} & {$60.6 \pm 0.1$} & {$70.2 \pm 0.1$} \\
{CR-PP} & {$52.1 \pm 0.1$} & {$62.4 \pm 0.1$} & {$71.9 \pm 0.1$} \\
{CR-VP} & {$52.9 \pm 0.2$} & {$62.8 \pm 0.1$} & {$72.8 \pm 0.1$} \\
\hline
\end{tabular}
}

\caption{\label{tab:f1results}Results for data augmentation experiments across different data set sizes. Top results for each data partition are marked in \textbf{bold}.}
\end{table}

All augmentation methods tested seem to improve performance in terms of F1 for the small training set (~50 sentences). When we look at the medium dataset, however, some methods such as CR-ADJP or CR-NP, start to have a negative impact compared to no augmentation settings. Even more augmentation strategies begin to show diminishing or negative effects on performance for the larger dataset (e.g., MR and CR-NP). This suggests that some of the augmented data might be detrimental for the model fine-tuning process. 

To understand how the augmented data may start to hurt the original model's performance, we consider one original sentence processed using CR-NP strategy. For example, ``Dr. Foutchner will arrange for an outpatient Holter monitor''.  In the case of the CR strategy, the augmentation algorithm draws an NP from another training sentence, resulting in ``Dr. Foutchner will arrange for a T2 signal change'' or  ``Dr. Foutchner will arrange for 10 beats''. These augmented sentences are grammatical, but they lack cohesion. This phenomenon may impact the model negatively.  Future work should explore strategies to control for this drift.  For instance, by fine-tuning a large-scale language model to perform masked language modeling on sentences where a portion of tokens are provided in terms of phrasal category (XP) or functional category (part of speech tag), we might hybridize syntax-driven transformations and instantiate syntactic templates using large-scale language models. 

We observed that among CR strategies, CR-NP performance seems to be worse compared to CR-VP or CR-PP, despite NP has the most occurrences in the training data. We suspect that the effectiveness of this strategy will heavily depend on the scope of the constituency tag. NPs are usually located low in the constituency tree, while VPs are usually located toward the top.

The augmentation strategies explored in this work can be further divided into two groups: strategies that produce new vocabularies and strategies that do not produce new vocabularies. SR and LM methods fall into augmentation that produce new vocabularies. SR uses the Penn Treebank \citep{marcus-etal-1993-building} to generate synonyms of replaced tokens. LM use its word embedding to guess the masked target token and may generate new words that do not exist in the training data. The other strategies (MR and CR) rely solely on the current training dataset. This phenomenon suggests augmentation strategies that produce new vocabularies seem to be more effective. This is plausible since new words will make the fine-tuned model more robust to unseen data. Although CR does not generate new words like the LM and SR methods, it still performs competitively in comparison. The delta between F1 scores produced by CR-VP and LM with our best hyperparameters for all dataset sizes are remarkably small at around 0 ~ 0.5 points.
The effect of simpler data augmentation strategies, SR and MR, seems to be diminishing as the data size increases; however, it is not the case with the LM and CR-VP strategies. They seem to perform well when more training data is available. 

\begin{table}[]
    \centering
    \scalebox{.65}{%
     \begin{tabular}{l|crr|crr|crr}
     & \multicolumn{3}{c|}{\textbf{S}} & \multicolumn{3}{c|}{\textbf{M}} & \multicolumn{3}{c}{\textbf{L}} \\ \hline
     & \textbf{5} & \multicolumn{1}{c}{\textbf{10}} & \multicolumn{1}{c|}{\textbf{20}} & \textbf{5} & \multicolumn{1}{c}{\textbf{10}} & \multicolumn{1}{c|}{\textbf{20}} & \textbf{5} & \multicolumn{1}{c}{\textbf{10}} & \multicolumn{1}{c}{\textbf{20}} \\ \cline{2-10} 
    \textbf{LM} & \multicolumn{1}{r}{50.6} & 51.5 & 50.6& \multicolumn{1}{r}{61.2} & 60.8 & 61.0 & \multicolumn{1}{r}{71.1} & 70.8 & 70.6 \\
    \textbf{CR-VP} & \multicolumn{1}{r}{50.5} & 52.0 & 52.3& \multicolumn{1}{r}{61.9} & 62.3 & 62.5 & \multicolumn{1}{r}{71.9} & 72.3 & 72.3 
    \end{tabular}

    }
    \caption{Comparison between CR-VP and LM augmentation. CR-VP holds more consistent performance across the number of generated sentences, while LM performance drops when the number of generated sentences is low.}
    \label{tab:robustness}
\end{table}

Looking at Table~\ref{tab:robustness}, the CR-VP augmentation strategy seems to show more consistent performance growth compared to the LM strategy. Whether it is 5, 10 or 20 sentences generated, CR-VP consistently trends upward as the number of augmented sentences increases (cf the instability of the LM). The average performance of the CR strategy shows an increased F1 as the number of synthetic sentences grows. In contrast, the average performance of the LM strategy is inconsistent and trending downward as the number of synthetic sentences increases. 

Lastly, the performance of the CR strategy will also be affected by the performance of constituency parser component itself. For one of our augmented examples, the original sentence ``She $[_{\textsc{VP}}$ had a workup by her neurologist$]$ and an MRI $[_{\textsc{VP}}$ call with any fevers , chills , increasing weakness... $]$'' was mutated into ``She $[_{\textsc{VP}}$ had a workup by her neurologist$]$ and an MRI $[_{\textsc{VP}}$ flare$]$''. Here, the word \textit{flare} was falsely predicted as a verb and thus erroneously predicted as a VP constituent, while the word \textit{flare} here should be a part of \textit{COPD flare} and classified as noun. 

\section{Conclusion and Future Work}
In this work, we examined data augmentation with a large-scale language model (LM) and constituency tree mutation (ST). We compared these augmentation methods with a baseline and previously proposed strategies for data augmentation: synonym replacement (SR) and mention replacement (MR). We found that SR performance is still most effective, by a small margin, but the performance degrades quickly as the data size increased. We have also observed that both LM and CR retained their performance throughout larger dataset sizes. We also showed that CR performance seems to be consistent in its improvement as the augmented dataset size increases, while the LM showed degrading performance with more augmented data. 

Future work should include improvements that hybridize the syntactic transformations with a large-scale language model. 
One possibility to increase the performance of the baseline language model is to train it to recognize phrase-level constituents and functional categories to understand more about constituency tags 
by first randomly swapping a few tokens with part of speech tags. For example, the original sentence is ``I take my medicine.'', then the pre-training sentence is ``I VB my medicine.'' and "I take my NN.". We hypothesize that this pre-training will improve the prediction performance of the baseline language model that we used for CR augmentation by attending to functional categories. Another possibility is to assign different weights to datapoints that inform the model how much to "trust" augmented data compared to gold data. This weight could be in the form of different learning rate.


\bibliography{main}
\bibliographystyle{acl_natbib}




\end{document}